\documentclass[letterpaper, 10 pt, conference]{ieeeconf}  

\usepackage{amssymb,balance}
\usepackage{amsmath}
\usepackage{algorithm}
\usepackage{algpseudocode}
\usepackage{graphicx}
\usepackage{subcaption}
\usepackage{xcolor}
\usepackage[sort,compress]{cite}
\usepackage{hyperref}
\hypersetup{
    colorlinks=true,
    linkcolor=blue,
    filecolor=magenta,      
    urlcolor=blue,
    }

\IEEEoverridecommandlockouts                              

\title{\LARGE \bf
Safe adaptation in multiagent competition
}

\author{Macheng Shen$^{1}$ and Jonathan P. How$^{2}$
\thanks{*This work is supported by ARL-DCIST (Cooperative Agreement Number W911NF-17-
2-0181).}
\thanks{$^{1}$Macheng Shen is with the Department of Mechanical Engineering,
        Massachusetts Institute of Technology, Cambridge, MA 02139, USA
        {\tt\small macshen@mit.edu}}%
\thanks{$^{2}$ Jonathan P. How is with the Department of Aeronautics and Astronautics, Massachusetts Institute of Technology, Cambridge, MA 02139, USA
        {\tt\small jhow@mit.edu}}%
}

\begin{document}

\maketitle
\thispagestyle{empty}
\pagestyle{empty}

\begin{abstract}

Achieving the capability of adapting to ever-changing environments is a critical step towards building fully autonomous robots that operate safely in complicated scenarios. In multiagent competitive scenarios, agents may have to adapt to new opponents with previously unseen behaviors by learning from the interaction experiences between the ego-agent and the opponent. However, this adaptation is susceptible to opponent exploitation. As the ego-agent updates its own behavior to exploit the opponent, its own behavior could become more exploitable as a result of overfitting to this specific opponent's behavior. To overcome this difficulty, we developed a safe adaptation approach in which the ego-agent is trained against a regularized opponent model, which effectively avoids overfitting and consequently improves the robustness of the ego-agent's policy. We evaluated our approach in the Mujoco domain with two competing agents. The experiment results suggest that our approach effectively achieves both adaptation to the specific opponent that the ego-agent is interacting with and maintaining low exploitability to other possible opponent exploitation. 

\end{abstract}

\section{INTRODUCTION}

One critical step towards building fully autonomous intelligent robots is to enable the capability of continual adaptation to new environments. In multiagent scenarios, besides the changing environment dynamics, agents must also adapt to the novel/evolving behaviors of other agents, which may not have been seen during training. There has been a lot of recent progress on fast adaptation to new tasks via meta-learning in both single-agent reinforcement learning (RL) \cite{rl_square, maml, meta_learning_1, meta_learning_2, reptile}, and multiagent reinforcement learning (MARL) \cite{meta_pg, lola, dongki}. During the meta-training phase, the agent meta-learns from tasks sampled from a task distribution how to quickly learn from a new task, which enables fast learning during the meta-testing phase. One important assumption within the meta-learning framework is that the task distribution is stationary \cite{continual_learning_review}. As a result, the tasks encountered during the meta-testing phase are sampled from the same distribution as those encountered during the meta-training phase. However, in multiagent competitive settings, assuming access to the task distribution (sampling from the unknown opponent agent's policy distribution) is often unrealistic. Furthermore, as the ego-agent adapts to the opponent, the opponent may also adapt to the ego-agent concurrently, leading to non-stationarity from the ego-agent's perspective \cite{non_stationary, dongki}. Achieving effective adaptation requires knowledge about the opponent's learning dynamics \cite{lola, dongki}, which is a very strong assumption in competitive scenarios.

In addition to the lack of knowledge about the opponent, another challenge of adaptation in competitive scenarios is to avoid being exploited by the opponent. In two-player zero-sum games, to exploit the opponent, the ego-agent has to deviate from the Nash-Equilibrium \cite{ne}, leading to increased exploitability. Previous works demonstrated search-based safe exploitation in extensive-form games \cite{extensive_form}, where the ego-agent updates its strategy via subgame-resolving leveraging a model of the opponent's strategy without substantially increasing the exploitability \cite{safe_exploit, safe_exploit_2, safe_exploit_3, Bayesian_exploit}. However, these approaches are specialized for extensive-form games with knowledge of the whole game tree, so it is unclear how to extend these approaches to handle more general multiagent settings, potentially with continuous dynamics, within the model-free reinforcement learning setting.

This paper focuses on safe adaptation in two-player zero-sum scenarios, where the ego-agent needs to update its policy based on the limited amount of interaction experience with an opponent agent. The goal of adaptation is to achieve high competitiveness (measured by cumulative reward) against this specific opponent that the ego-agent is interacting with, while the safety requirement is that the ego-agent must also maintain high competitiveness against any other possible opponent (with either a stationary or evolving policy) during the whole adaptation phase. We investigate this problem under the framework of Markov game \cite{markov_game} and multiagent reinforcement learning in a model-free setting without explicit assumptions on the state or action space.
As such, the main contributions of this paper are:
\begin{enumerate}
    \item We present a novel Bayesian formulation of the safe adaptation problem within the MARL framework, which bridges the connection between robust MARL and safe adaptation. 
    \item We proposed an optimization objective for modeling the opponent, with a behavior cloning term for adaptation and a novel ensemble-regularization term to achieve low exploitability, which is derived from the Bayesian formulation.
    \item We demonstrated that our approach achieves adaptation by learning from a limited amount of interaction experience with the opponent while maintaining low exploitability against a second opponent that actively co-adapts to exploit the ego-agent.
\end{enumerate}

\section{Preliminaries}

\subsection{Markov Games}

A Markov game for $N$ agents is defined by a set of states $\mathcal{S}$ describing the possible configurations of all agents, a set of actions $\mathcal{A}_1,\ldots,\mathcal{A}_N$, and a set of observations $\mathcal{O}_1,\ldots,\mathcal{O}_N$ for each agent. Each agent has a stochastic policy $\pi_i: \mathcal{O}_{i} \times \mathcal{A}_{i} \mapsto[0,1]$, and a reward function $r_{i}: \mathcal{S} \times \mathcal{A}_{i} \mapsto \mathbb{R}$.

\subsection{Multiagent reinforcement learning}
The objective of each agent is to maximize its own cumulative reward $R_{i}=\sum_{t=0}^{T} \gamma^{t} r_{i}^{t}$ with discount factor $\gamma$ and time horizon $T$ \cite{maddpg}. As a result, the learning problem is formulated as finding a joint policy $\boldsymbol{\pi} = \{\pi_i\}^{i=1:N}$, where each policy maximizes its own reward, 
\begin{equation}
    J_i = \mathbb{E}_{s \sim p^{\boldsymbol{\pi}}, a_i \sim \pi_{i}, \boldsymbol{a}_{-i} \sim \boldsymbol{\pi}_{-i}} \left[R_{i}(s, \boldsymbol{a}) \right],
\label{marl_with_single_policy}
\end{equation}
with $p^{\boldsymbol{\pi}}$ being the the transition dynamics induced by the joint policy $\boldsymbol{\pi}$ and the subscript $-i$ denotes the set $\{j|j \neq i, j=1,2,\ldots,N \}$.

\subsection{MARL with policy distribution}
Various empirical studies \cite{maddpg, alphaGo, capture_the_flag, alphastar, psro, emergent, robust_adversarial_population} and a recent theoretical study \cite{spinning} suggest a more general formulation of MARL in which each agent samples its policy from a policy distribution. Therefore, we consider the following objective function that learns a distribution of policies for each agent,
\begin{equation}
    J_i = \mathbb{E}_{s \sim p^{\boldsymbol{\pi}},
    \boldsymbol{a} \sim \boldsymbol{\pi},
    \boldsymbol{\pi} \sim \mathcal{P}(\boldsymbol{\Pi}),
    } \left[R_{i}(s, \boldsymbol{a}) \right],
\label{marl_with_policy_distribution}
\end{equation}
where $\mathcal{P}$ is a joint distribution over the joint policy space $\boldsymbol{\Pi} = \Pi_1 \times \Pi_2 \ldots \times \Pi_N$. Each agent is learning its own policy distribution $\Pi_i$ to optimize its objective $J_i$ subject to the joint distribution $\boldsymbol{\Pi}$.

Note that the feasibility set of Eq.~\ref{marl_with_policy_distribution} contains that of Eq.~\ref{marl_with_single_policy}, which is analogous to the relationship between a mixed-strategy Nash Equilibrium and a pure-strategy Nash Equilibrium \cite{nash_equilibrium}. This relationship also suggests that Eq.~\ref{marl_with_policy_distribution} is a more appropriate learning objective than Eq.~\ref{marl_with_single_policy}.

\section{Our approach}

As we focus on two-player zero-sum games, we rewrite Eq.~\ref{marl_with_policy_distribution} from the ego-agent's perspective as:
\begin{equation}
    J_{\text{ego}} = \mathbb{E}_{\pi_{\text {ego}} \sim p\left(\mathbf{\Pi}_{\text {ego}}\right), \pi_{\text {oppo}} \sim p\left(\mathbf{\Pi}_{\text {oppo}}\right)}\left[\mathbb{E}_{s \sim p^{\boldsymbol{\pi}},
    \boldsymbol{a} \sim \boldsymbol{\pi}}\left[R_{\text {ego}}(s, \boldsymbol{a})\right]\right],
    \label{ego_view_objective}
\end{equation}
where the ego-agent optimizes its policy distribution 
$\mathbf{\Pi}_{\text {ego}}$, subject to a given opponent policy distribution $\mathbf{\Pi}_{\text {oppo}}$. Since the optimal $\mathbf{\Pi}_{\text {ego}}$ with respect to Eq.~\ref{ego_view_objective} depends on the opponent policy distribution $\mathbf{\Pi}_{\text {oppo}}$, determining $\mathbf{\Pi}_{\text {oppo}}$ is critical. Here, we discuss some common choices for the opponent policy distribution $\mathbf{\Pi}_{\text {oppo}}$:
\begin{enumerate}
    \item \textbf{Oracle policy distribution}: Suppose we know the true policy distribution of the opponent, we can optimize the ego-agent policy distribution against the opponent policy distribution to obtain $\mathbf{\Pi}_{\text {ego}}^{\text{oracle}}$.     
    However, there are two problems with this approach: 1) Feasibility: In competitive scenarios, it is unlikely to get access to the policy of the opponent. 2) Robustness: As the ego-agent over-fits its policy to the opponent, the resulting $\mathbf{\Pi}_{\text {ego}}^{\text{oracle}}$ may not perform well (or even poorly as shown in \cite{adv_attack}) against an adversarial opponent that is trained against $\mathbf{\Pi}_{\text {ego}}^{\text{oracle}}$ to exploit its weakness.
    
    \item \textbf{Learned opponent policy distribution \cite{oppo_modeling_drl, som, policy_embedding, vae_oppo_modeling}}: As the ego-agent interacts with the opponent, the ego-agent can learn an internal model of the opponent policy $\mathbf{\Pi}_{\text {oppo}}^{\text{model}}$ distribution from the interaction experience as an approximation of the true opponent policy distribution. Many previous works \cite{oppo_modeling_drl, som, policy_embedding}, assume access to the opponent's observation and action for this model learning, which is not a strong assumption in robotic domains with full-observability over the state space since the opponent's observation and action can be deducted from the state observation. Besides, \cite{vae_oppo_modeling} also demonstrates the possibility of learning an opponent model from the ego-agent's observation alone via variational inference over a hidden space that models the opponent's private information. However, this opponent modeling approach also suffers from the robustness problem mentioned earlier. 
    
    \item \textbf{Nash Equilibrium policy distribution}: Another way to model the opponent is to solve for the Nash Equilibrium policy distribution $\mathbf{\Pi}_{\text {oppo}}^{\text{nash}}$ (or equivalently, the minimax solution \cite{minimax} in two-player zero-sum games \cite{nash_equilibrium}), with no prior knowledge of the opponent's policy distribution. The corresponding optimal policy distribution for the ego-agent is also the Nash Equilibrium $\mathbf{\Pi}_{\text {ego}}^{\text{nash}}$, which is the least exploitable policy. However, this approach does not attempt to adapt to the opponent that the ego-agent is interacting with, leading to sub-optimal performance against an opponent with exploitability.
\end{enumerate}

We argue that a better approach to modeling the opponent's policy distribution should leverage the available interaction experience for adaptation as well as stay close to the equilibrium distribution for robustness against adversarial exploitation. We derive this approach by rephrasing the sub-problem within Eq.~\ref{ego_view_objective} of modeling the opponent policy distribution $\pi_{\text {oppo}} \sim p\left(\mathbf{\Pi}_{\text {oppo}}\right)$ into a Bayesian inference problem over the space of policy distributions given the interaction experience $\mathcal{D}$ between the ego-agent and the opponent:
\begin{equation}
    \begin{aligned}
            \pi_{\text {oppo}} \sim p\left(\mathbf{\Pi}_{\text {oppo}}|\mathcal{D}\right) & \propto
            p\left(\mathbf{\Pi}_{\text {oppo}}|\mathcal{\emptyset}
            \right)
            \times p\left(\mathcal{D}|\mathbf{\Pi}_{\text {oppo}}\right) ,\\
            &
            = p^{\text{NE}}\left(\mathbf{\Pi}_{\text {oppo}} \right)\times
            p\left(\mathcal{D}|\mathbf{\Pi}_{\text {oppo}}\right),
    \end{aligned}
    \label{posterior_policy_distribution}
\end{equation}
where $\emptyset$ denotes the empty set, so $p^{\text{prior}}\left(\mathbf{\Pi}_{\text {oppo}}\right) = p\left(\mathbf{\Pi}_{\text {oppo}}|\mathcal{\emptyset} \right)$ is the prior distribution over the opponent's policy space before obtaining any interaction experience. We argue that the Nash Equilibrium policy distribution is a sensible choice for this prior, i.e. $p^{\text{prior}}\left(\mathbf{\Pi}_{\text {oppo}}\right) = p^{\text{NE}}\left(\mathbf{\Pi}_{\text {oppo}}\right)$, since with no information about the opponent, the best choice is to minimize the ego-agent's exploitability. As the ego-agent receives more interaction experience with the opponent, the posterior distribution $p\left(\mathbf{\Pi}_{\text {oppo}}|\mathcal{D}\right)$ is updated through the likelihood term $p\left(\mathcal{D}|\mathbf{\Pi}_{\text {oppo}}\right)$ while regularized by the prior term, which ensures adaptation to the opponent while maintaining low exploitability. 

However, the posterior inference problem Eq.~\ref{posterior_policy_distribution} is challenging for two reasons: 1) Solving for the Nash Equilibrium policy distribution $p^{\text{NE}}\left(\mathbf{\Pi}_{\text {oppo}}\right)$ is a challenging problem; 2) representing and parameterizing policy distribution is challenging. Therefore, we apply the following two approximations,

\begin{enumerate}
    \item We approximate the Nash Equilibrium policy distribution via an ensemble of policies generated via Alg. \ref{alg:ensemble_training}, which has been shown in \cite{maddpg, adv_attack, emergent} to produce robust agent behaviors that are much less exploitable than policies generated without ensembling.
    \item Instead of modeling the posterior distribution, we seek for a single opponent policy that approximates the maximum a posteriori probability (MAP) estimate of Eq.~\ref{posterior_policy_distribution}.
\end{enumerate}
With these two approximations, we propose an alternative formulation for the estimated opponent policy as the optimization problem in Eq.~\ref{alternative_formulation}.
\begin{equation}
    \begin{aligned}
        \hat{\pi}_{\text {oppo}}=\arg\min _{\pi} [\mathbb{D} (\mathbf{\Pi}_{\text {oppo}}^{\text{ensemble}} \mid \pi) & + \lambda_{1} \mathbb{L}_{\text{likelihood}}(\mathcal{D} \mid \pi) \\ & + \lambda_{2} \mathbb{L}_{\mathrm{RL}}(\mathbf{\Pi}^{\text{ensemble}}_{\text {ego}}, \pi)],
    \end{aligned}
    \label{alternative_formulation}
\end{equation}
where $\lambda_1$ and $\lambda_2$ are hyper-parameters. The first term $\mathbb{D} (\mathbf{\Pi}_{\text {oppo}}^{\text{ensemble}} \mid \pi)$ denotes a distance metric between the opponent policy ensemble generated via Alg. \ref{alg:ensemble_training} and the estimated opponent model, which regularizes the opponent policy to stay close to the robust ensemble policy distribution. This term corresponds to the prior term $p^{\text{NE}}\left(\mathbf{\Pi}_{\text {oppo}}\right)$ in Eq.~\ref{posterior_policy_distribution}. The second term $\mathbb{L}_{\text{likelihood}}(\mathcal{D} \mid \pi)$ is the log-likelihood of observing the interaction experience $\mathcal{D}$ given the opponent policy, which corresponds to the likelihood term $p\left(\mathcal{D}|\mathbf{\Pi}_{\text {oppo}}\right)$ in Eq.~\ref{posterior_policy_distribution}. The third term $\mathbb{L}_{\mathrm{RL}}(\mathbf{\Pi}^{\text{ensemble}}_{\text {ego}}, \pi)]$ is the reinforcement learning loss that optimizes the opponent policy against the ego-agent's policy ensemble,
\begin{equation}
    \begin{aligned}
        & \mathbb{L}_{\mathrm{RL}}(\mathbf{\Pi}^{\text{ensemble}}_{\text {ego}}, \pi) \\
        & = - \mathbb{E}_{\pi_{\text {ego}} \sim p\left(\mathbf{\Pi}^{\text{ensemble}}_{\text {ego}}\right), \pi_{\text {oppo}} = \pi} \left[\mathbb{E}_{s \sim p^{\boldsymbol{\pi}},
\boldsymbol{a} \sim \boldsymbol{\pi}}\left[R_{\text {oppo}}(s, \boldsymbol{a})\right]\right],
    \end{aligned}
    \label{RL_loss}
\end{equation}
where the minus sign ensures that minimizing this loss results in maximization of the reward. This term does not correspond to any term in Eq.~\ref{posterior_policy_distribution}. Intuitively, this term enables the opponent policy to continually evolve, as we update the ego-agent's policy ensemble to adapt to the opponent via optimizing its reinforcement learning objective in Eq.~\ref{ego_agent_update},
\begin{equation}
    J_{\text{ego}} = \mathbb{E}_{\pi_{\text {ego}} \sim p\left(\mathbf{\Pi}^{\text{ensemble}}_{\text {ego}}\right), \pi_{\text {oppo}} = \hat{\pi}_{\text {oppo}}}\left[\mathbb{E}_{s \sim p^{\boldsymbol{\pi}},
    \boldsymbol{a} \sim \boldsymbol{\pi}}\left[R_{\text {ego}}(s, \boldsymbol{a})\right]\right].
    \label{ego_agent_update}
\end{equation}
We include $\mathbb{L}_{\mathrm{RL}}$ because we found that although this term does not make much difference when adapting to a stationary opponent, it could be critical for achieving high competitiveness against an evolving opponent as we show later in the experiment section. 

\begin{algorithm}
    \caption{Ensemble training}\label{alg:ensemble_training}
        \begin{algorithmic}[1]
            \Require Ensemble size $N$, number of training iterations $K$
            \State Randomly initialize policy ensembles: $\mathbf{\Pi}^{\text{ensemble}}_{\text{ego}} = \{\pi^i_{\text{ego}}\}_{i=1:N}$, $\mathbf{\Pi}^{\text{ensemble}}_{\text{oppo}} = \{\pi^i_{\text{oppo}}\}_{i=1:N}$
            \For{k = 1:$K$}
                \State Randomly sample policy index: $j\sim \{1, \ldots,N\}, l \sim \{1,\ldots,N\}$
                \State Environment\_rollout($\pi^j_{\text{ego}}$, $\pi^l_{\text{oppo}}$)
                \State Update $\pi^j_{\text{ego}}$ and $\pi^l_{\text{oppo}}$ to optimize objective Eq.~\ref{marl_with_policy_distribution}
            \EndFor
            \State \Return $\mathbf{\Pi}^{\text{ensemble}}_{\text{ego}}$ and $\mathbf{\Pi}^{\text{ensemble}}_{\text{oppo}}$
        \end{algorithmic}
\end{algorithm}

\begin{algorithm}
    \caption{Safe adaptation}\label{alg:safe_adaptation}
        \begin{algorithmic}[1]
            \Require Policy ensembles $\mathbf{\Pi}^{\text{ensemble}}_{\text{ego}}$ and $\mathbf{\Pi}^{\text{ensemble}}_{\text{oppo}}$, interaction experience $\mathcal{D}$, number of iterations $K$
            \State freeze $\mathbf{\Pi}^{\text{ensemble}}_{\text{oppo}}$
            \State Initialize $\hat{\pi}^0_{\text {oppo}}$ as random policy
            \State $\mathbf{\Pi}^{\text{ensemble,0}}_{\text{oppo}} \gets \mathbf{\Pi}^{\text{ensemble}}_{\text{oppo}}$
            \For{k = 1:$K$}
                \State $\hat{\pi}^k_{\text {oppo}} \gets  update\_opponent(\hat{\pi}^{k-1}_{\text {oppo}}, \mathbf{\Pi}^{\text{ensemble}}_{\text{oppo}},$ 
                \State $\mathbf{\Pi}^{\text{ensemble},k-1}_{\text{ego}}, \mathcal{D}$) \Comment{one gradient step of Eq.~\ref{alternative_formulation}}
                \State $\mathbf{\Pi}^{\text{ensemble},k}_{\text{ego}} \gets update\_ego\_agent(\hat{\pi}^k_{\text {oppo}}, \mathbf{\Pi}^{\text{ensemble}, k-1}_{\text{ego}})$ \Comment{one gradient step of Eq.~\ref{ego_agent_update}}
            \EndFor
            \State \Return $\mathbf{\Pi}^{\text{ensemble},K}_{\text{ego}}$
        \end{algorithmic}
\end{algorithm}

Now we discuss our choice for the first two terms in Eq.~\ref{alternative_formulation}. The first term $\mathbb{D} (\mathbf{\Pi}_{\text {oppo}}^{\text{ensemble}} \mid \pi)$ measures the discrepancy between the policy ensemble $\mathbf{\Pi}_{\text {oppo}}^{\text{ensemble}}$ and our estimated opponent policy $\pi$. There are several closed-form metrics to measure the discrepancy between two policies, including KL-divergence discrepancy \cite{kl_divergence_policy_discrepancy}, total variation distance \cite{total_variation_distance} and maximum mean discrepancy \cite{max_mean_discrepancy}. However, it is unclear how to select one metric over another given a specific application domain, and whether the selected metric can optimally discriminate between two policies. To resolve this ambiguity and achieving optimal discriminative power, we choose to learn the discrepancy metric via adversarial learning following the paradigm of generative adversarial imitation learning (GAIL) \cite{GAIL}, where we train a discriminator $D_w(o, a): \mathcal{O}_{\text{oppo}}\times \mathcal{A}_{\text{oppo}} \rightarrow [0, 1]$ to minimize the following discrimination loss,
\begin{equation}
    \mathbb{E}_{\tau_{\pi}}[\log D_w(o, a)] + \mathbb{E}_{\tau_{\mathbf{\Pi}_{\text {oppo}}^{\text{ensemble}}}}[\log (1 - D_w(o, a))],
\end{equation}
such that
\begin{equation}
    \mathbb{D} (\mathbf{\Pi}_{\text {oppo}}^{\text{ensemble}} \mid \pi) = - \mathbb{E}_{\tau_{\pi}}[\log D_w(o, a)],
    \label{gail_loss}
\end{equation}
where the shorthand notation $\tau_{(\cdot)}$ denotes the trajectory distribution when the ego-agent follows policy $\pi_{\text {ego}} \sim p\left(\mathbf{\Pi}_{\text {ego}}\right)$, while the opponent follows policy $(\cdot)$. This loss function is minimized by maximizing an imitation reward $r_{\text{imit}} = \log D_w(o, a)$.

The second term in Eq.~\ref{alternative_formulation} is the log-likelihood of observing the experience $\mathcal{D}$ given the opponent policy $\pi$. Since the opponent policy can only affect the probability of the opponent's taken action, this term can be reduced to behavior cloning loss,
\begin{equation}
    p\left(\mathcal{D}\mid \pi \right) = 
    \sum_{(a,o) \in \mathcal{D}} \log \pi(a \mid o).
\end{equation}
In practice, we use mini-batch to calculate the gradient of this loss.

\section{Results and discussion}
\subsection{Experiment setting}
\begin{figure}[t]
    \centering
    \includegraphics[width=\columnwidth]{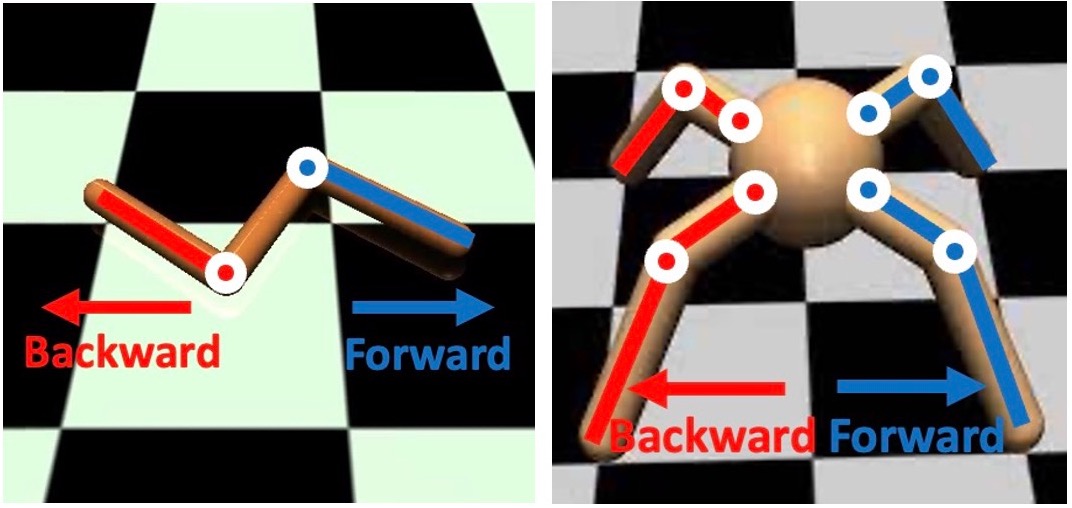}
    \caption{Mujoco environments (left: swimmer; right: ant), where the blue part of the body is controlled by the ego-agent and the red part of the body is controlled by the opponent agent. The white circles are the joints where agents can apply torques on. The ego-agent is rewarded for moving forward while the opponent agent is rewarded for moving backward.}
    \label{fig:envs}
\end{figure}

We evaluate our safe adaptation approach on the Multiagent Mujoco domain \cite{ma_mujoco}, where each robot is decomposed into parts that are controlled by individual agents as illustrated in Fig.~\ref{fig:envs}. We use a zero-sum reward where the ego-agent tries to maximize the reward for moving forward and the opponent agent tries to minimize this reward. 

To evaluate the capability of safe adaptation to a previously unseen opponent, we describe the following procedure to set up the evaluation:

\textbf{Off-line training phase}: 
        \begin{enumerate}
            \item Alternating for $K_1$ iterations, between one-step ($K=1$) policy ensemble training of size $N=5$ for both agents, $\mathbf{\Pi}^{\text{ensemble}}_{\text{ego}}, \mathbf{\Pi}^{\text{ensemble}}_{\text{oppo}}$ via Alg. \ref{alg:ensemble_training} and one-step training of exploiter opponent $\pi^{\text{exp}}_{\text{oppo}}$ via Alg. \ref{alg:opponent_exploiter}.
            \item Freeze $\mathbf{\Pi}^{\text{ensemble}}_{\text{ego}}, \mathbf{\Pi}^{\text{ensemble}}_{\text{oppo}}$, and training exploiter opponent $\pi^{\text{exp}}_{\text{oppo}}$ against $\mathbf{\Pi}^{\text{ensemble}}_{\text{ego}}$ for an additional $K_2$ iterations.
        \end{enumerate}
            
\textbf{On-line adaptation phase} (for $\mathbf{\Pi}^{\text{ensemble}}_{\text{ego}}$ to adapt to $\pi^{\text{exp}}_{\text{oppo}}$): 
    \begin{enumerate}
        \item Freeze $\pi^{\text{exp}}_{\text{oppo}}$, $\mathbf{\Pi}^{\text{ensemble}}_{\text{oppo}}$, and $\mathbf{\Pi}^{\text{ensemble}}_{\text{ego}}$.
        \item Collect interaction experience $\mathcal{D}$ between $\mathbf{\Pi}^{\text{ensemble}}_{\text{ego}}$ and $\pi^{\text{exp}}_{\text{oppo}}$.
        \item Unfreeze $\mathbf{\Pi}^{\text{ensemble}}_{\text{ego}}$.
        \item Initialize second exploiter opponent ${\pi^*}^{\text{exp}}_{\text{oppo}}$ from $\pi^{\text{exp}}_{\text{oppo}}$.
        \item Alternating for $K_3$ iterations, between one-step safe adaptation for $\mathbf{\Pi}^{\text{ensemble}}_{\text{ego}}$ to adapt to $\pi^{\text{exp}}_{\text{oppo}}$ via Alg. \ref{alg:safe_adaptation} and one-step training of the second exploiter opponent ${\pi^*}^{\text{exp}}_{\text{oppo}}$ to exploit $\mathbf{\Pi}^{\text{ensemble}}_{\text{ego}}$ via Alg. \ref{alg:opponent_exploiter}.
    \end{enumerate}

During the off-line training phase, we alternate between ensemble training $\mathbf{\Pi}^{\text{ensemble}}_{\text{ego}}$ and $\mathbf{\Pi}^{\text{ensemble}}_{\text{oppo}}$, and training of exploiter opponent $\pi^{\text{exp}}_{\text{oppo}}$ to mitigate the well-known problem of training imbalance \cite{emergent, gan} in competitive/adversarial training, so that the exploiter opponent can always catch up with the ego-agent before the ego-agent becomes too strong. The additional training of the exploiter opponent ensures that the exploiter is sufficiently trained to exploit the ego-agent, which motivates the ego-agent to adapt to this exploiter in the adaptation phase. During the whole off-line training phase, the ego-agent does not collect experience against the exploiter opponent $\pi^{\text{exp}}_{\text{oppo}}$. As a result, this exploiter is a previously unseen opponent in the ego-agent's perspective.

During the on-line adaptation phase, the ego-agent policy ensemble $\mathbf{\Pi}^{\text{ensemble}}_{\text{ego}}$ adapts to the exploiter opponent $\pi^{\text{exp}}_{\text{oppo}}$ given a fixed size interaction experience $\mathcal{D}$, with regularization from $\mathbf{\Pi}^{\text{ensemble}}_{\text{oppo}}$. Concurrently, the second exploiter ${\pi^*}^{\text{exp}}_{\text{oppo}}$ is trained to exploit the $\mathbf{\Pi}^{\text{ensemble}}_{\text{ego}}$. As a result, the reward against the first exploiter opponent $\pi^{\text{exp}}_{\text{oppo}}$ during the adaptation phase measures the capability of adaptation against a stationary opponent, while the reward against the second exploiter opponent ${\pi^*}^{\text{exp}}_{\text{oppo}}$ measures robustness/safety against an evolving adversarial exploiter.

\begin{algorithm}[t]
    \caption{Train exploiter opponent}\label{alg:opponent_exploiter}
        \begin{algorithmic}[1]
            \Require Ego-agent ensemble $\mathbf{\Pi}^{\text{ensemble}}_{\text{ego}}$, exploiter opponent policy $\pi^{\text{exp}}_{\text{oppo}}$, number of training iterations $K$
            \State freeze $\mathbf{\Pi}^{\text{ensemble}}_{\text{ego}}$
            \For{k = 1:$K$}
                \State Train  $\pi^{\text{exp}}_{\text{oppo}}$ against $\mathbf{\Pi}^{\text{ensemble}}_{\text{ego}}$ by gradient decent on $\mathbb{L}_{\mathrm{RL}}(\mathbf{\Pi}^{\text{ensemble}}_{\text {ego}}, \pi^{\text{exp}}_{\text{oppo}})$
            \EndFor
            \State Unfreeze $\mathbf{\Pi}^{\text{ensemble}}_{\text{ego}}$
            \State \Return $\pi^{\text{exp}}_{\text{oppo}}$
        \end{algorithmic}
\end{algorithm}

\begin{figure*}[thbp]
     \centering
     \begin{subfigure}[b]{0.49\textwidth}
         \centering
         \includegraphics[width=\textwidth,trim= 0 5 0 20,clip]{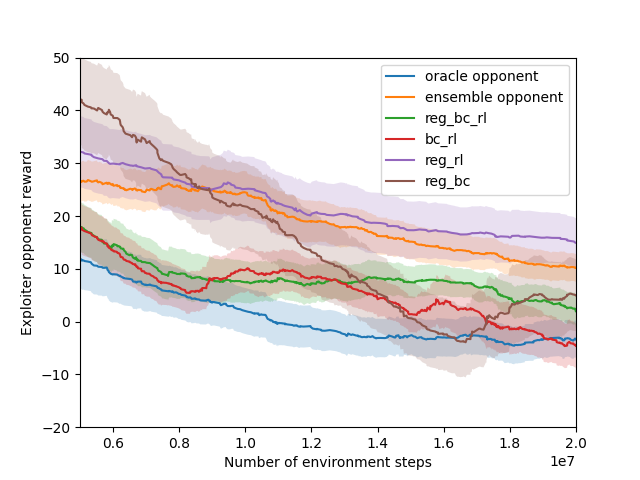}
         \caption{Swimmer adaptation against first exploiter}
         \label{fig:swimmer_first_exploiter}
     \end{subfigure}
     \hfill
     \begin{subfigure}[b]{0.49\textwidth}
         \centering
         \includegraphics[width=\textwidth,trim= 0 5 0 20,clip]{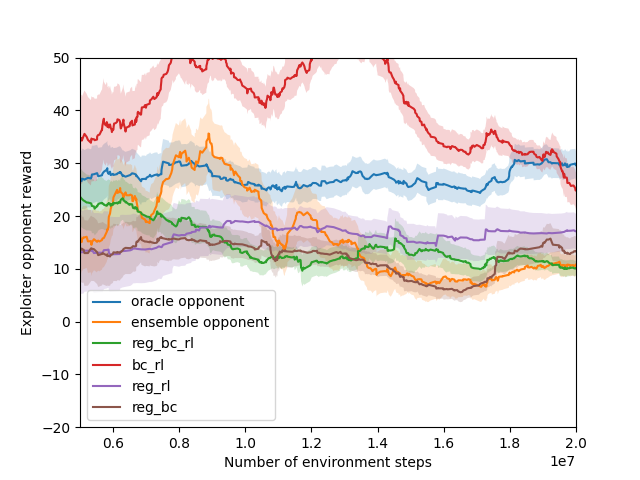}
         \caption{Swimmer adaptation against second exploiter}
         \label{fig:swimmer_second_exploiter}
     \end{subfigure}
     \hfill
\vspace*{-0.025in}
     \begin{subfigure}[b]{0.49\textwidth}
         \centering
         \includegraphics[width=\textwidth,trim= 0 5 0 30,clip]{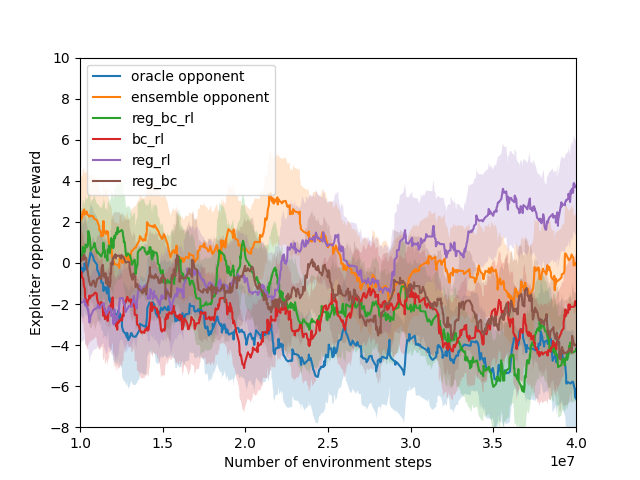}
         \caption{Ant adaptation against first exploiter}
         \label{fig:ant_first_exploiter}
     \end{subfigure}
     \hfill
     \begin{subfigure}[b]{0.49\textwidth}
         \centering
         \includegraphics[width=\textwidth,trim= 0 5 0 30,clip]{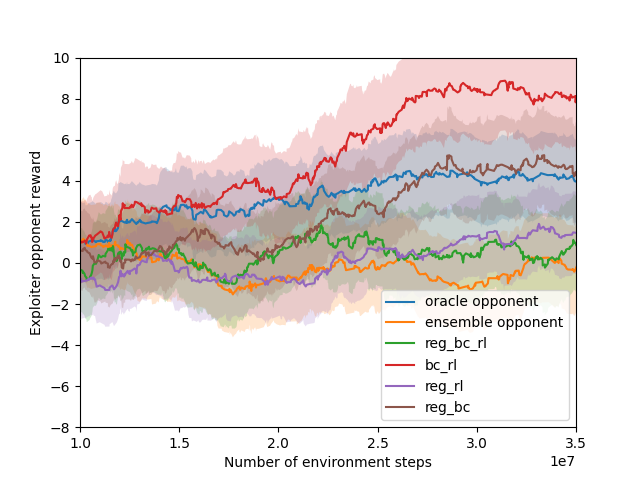}
         \caption{Ant adaptation against second exploiter}
         \label{fig:ant_second_exploiter}
     \end{subfigure}
\vspace*{-0.025in}

        \caption{Exploiter opponent rewards during the adaptation phase. Lower opponent reward against the first (\ref{fig:swimmer_first_exploiter}, \ref{fig:ant_first_exploiter}) and the second (\ref{fig:swimmer_second_exploiter}, \ref{fig:ant_second_exploiter}) exploiter indicates better adaptation and safety of the ego-agent policy, respectively. Our proposed approach (reg\_bc\_rl) achieves near-optimal performance  (w.r.t. the references) at both adaptation and safety, while the references (oracle opponent and ensemble opponent) achieve good performance only at one of these two metrics, but poor performance at the other.}
        \label{fig:exploiter_opponent_rewards}
\end{figure*}

We use stochastic policy with Gaussian distribution, and two fully-connected hidden layers, each with 128 hidden units followed by ReLU activate layer, as the policy and critic network architecture. We use a pytorch implementation\footnote{\href{https://github.com/p-christ/Deep-Reinforcement-Learning-Algorithms-with-PyTorch}{https://github.com/p-christ/Deep-Reinforcement-Learning-Algorithms-with-PyTorch}}
of Soft Actor-Critic \cite{sac} (SAC) with dual critic networks and automatic tuning of the entropy parameter to train the ensemble networks and the two exploiter opponents in both the off-line training phase and the on-line adaptation phase. We use a replay buffer size of one million for the off-line training phase but reduce that to half a million for the on-line adaptation phase to save memory. Our adaptation implementation is modified from the GAIL implementation in PyTorch-RL\footnote{\href{https://github.com/Khrylx/PyTorch-RL}{https://github.com/Khrylx/PyTorch-RL}} with Proximal Policy Optimization \cite{ppo} (PPO) for training $\hat{\pi}_{\text {oppo}}$, with PPO rollout batch size of 1000, mini-batch size of 128, and 10 gradient updates per PPO batch. We use a learning rate 0.001 for both training and adaptation. 
In both environments, the number of steps per episode is fixed at 500. Each agent can observe the joints/bodies position and velocity of its own and its opponent's, so the agents have full observability. 

In the off-line training phase, we use $K_1 = 10000$ iterations (episodes), and $K_2 = 5000$ iterations the swimmer environment and $K_2 = 2000$ for the ant environment. In the on-line adaptation phase, we use different $K_3$ for the two environments until the average rewards become steady. We collect 10 episodes of interaction experience for adaptation, which corresponds to $|\mathcal{D}|=5000$ environment steps.  We tune the hyper-parameters $\lambda_1$ and $\lambda_2$ independently for each environment. We select from the following values: $\{0.1, 0.5, 1.0, 5.0\}$, and manually choose the best one by looking at the adaptation rewards against both of the two exploiter opponents. The selected hyper-parameters are: $\lambda_1 = 1.0, \lambda_2 = 1.0$ for the swimmer environment, and $\lambda_1 = 0.1, \lambda_2 = 1.0$ for the ant environment. 

\subsection{Results}
We show the exploiter opponent rewards during the adaptation phase in Fig.~\ref{fig:exploiter_opponent_rewards}, which includes the following settings: 1) oracle opponent, where the ego-agent is trained against the first exploiter directly. This setting is unrealistic since the first exploiter's policy is unknown to the ego-agent; 2) ensemble opponent, where the ego-agent is trained against the opponent ensemble policy generated from the off-line training phase; 3) reg\_bc\_rl: our proposed approach including all the three terms in Eq.~\ref{alternative_formulation}; 4) bc\_rl: ablation of our approach without the ensemble regularization term; 5) reg\_rl: ablation of our approach without the behavior cloning term; 6) reg\_bc: ablation of our approach without the RL term. 

Fig.~\ref{fig:swimmer_first_exploiter} shows that the oracle opponent setting (reference) achieves the best adaptation against the first exploiter opponent, while all the settings with the behavior cloning loss achieve comparable adaptation performance as the reference. For all the settings, the opponent reward decreases due to the fact that the opponent policy is fixed but the ego-agent policy is updating. However, there remains a gap between those settings without interaction experience with the first exploiter opponent (ensemble opponent and reg\_rl) and the other settings.   

Fig.~\ref{fig:swimmer_second_exploiter} shows that those settings without the ensemble regularization term (oracle opponent and bc\_rl) are unable to achieve robustness against the second exploiter opponent which actively exploits the ego-agent's policy. Combining Fig.~\ref{fig:swimmer_first_exploiter} and \ref{fig:swimmer_second_exploiter}, we can conclude that our approach (reg\_bc\_rl) strikes a better trade-off between adaptation and safety, compared with the two reference approaches (oracle opponent: good adaptation but poor robustness; ensemble opponent: good robustness but poor adaptation).

Fig.~\ref{fig:ant_first_exploiter} and \ref{fig:ant_second_exploiter} show consistent results as those in Fig.~\ref{fig:swimmer_first_exploiter} and \ref{fig:swimmer_second_exploiter}: the ego-agent's policies that are adapting to the first exploiter opponent are also more susceptible to be exploited by the second exploiter opponent. Besides, Fig.~\ref{fig:ant_second_exploiter} shows that the setting without the RL loss (reg\_bc) is also exploited after the second exploiter is sufficiently trained. Our conjectured reason for this observation is that the RL term enables the estimated opponent policy $\hat{\pi}_{\text{oppo}}$ to discover the weakness of the ego-agent's policy, which also helps reduce the ego-agent's exploitability because the ego-agent is trained against the estimated opponent policy.

To quantitatively measure the adaptation and robustness of different settings, we calculate the area under curve (AUC) metrics of both the first exploiter's (measures adaptation) and the second exploiter's (measures robustness) reward curves normalized by the two reference settings (oracle opponent and ensemble opponent), as shown in Table~\ref{table:auc_exploiter_reward}. We calculate the AUC using reward curves from steps 1e7 to 2e7 for the swimmer environment and from steps 1e7 to 3.5e7 for the ant environment. These results are consistent with our hypothesis that learning from the interaction experience with the opponent enables adaptation, but without regularization from the ensemble policy, this adaptation could be highly exploitable. The regularization term is effective for achieving safe adaptation. As a result, our approach achieves the best overall metric which combines adaptation and robustness.

\begin{table}[t]
    \centering
    \caption{Adaptation and robustness metrics of different settings, where the oracle opponent and the ensemble opponent settings are taken as references. The best two settings are highlighted.}
    
    \begin{tabular}{lcccr}
        \textbf{Swimmer} & & &\\
        \hline
        Settings/Metrics & Adaptation & Robustness & Overall (A+R)\\
        \hline
        \textit{Oracle opponent} & 1.0 & 0.0 & 1.0\\
        \textit{Ensemble opponent} & 0.0 & 1.0 & 1.0\\
        Reg\_bc\_rl & 0.50 & \textbf{1.01} & \textbf{1.51}\\
        Bc\_rl & \textbf{0.67} & -0.92 & -0.25\\
        Reg\_rl & -0.16 & 0.69 & 0.53\\
        Reg\_bc & \textbf{0.52} & \textbf{1.07} & \textbf{1.59}\\
        \hline
    \end{tabular}
    
    \begin{tabular}{lcccr}
        \\
        \textbf{Ant} & & \\
        \hline
        Settings/Metrics & Adaptation & Robustness & Overall (A+R)\\
        \hline
        \textit{Oracle opponent} & 1.0 & 0.0 & 1.0\\
        \textit{Ensemble opponent} & 0.0 & 1.0 & 1.0\\
        Reg\_bc\_rl & \textbf{0.60} & \textbf{0.79} & \textbf{1.39}\\
        Bc\_rl & \textbf{0.76} & -0.60 & 0.16\\
        Reg\_rl & 0.06 & \textbf{0.89} & \textbf{0.95}\\
        Reg\_bc & 0.48 & 0.25 & 0.73\\
        \hline
    \end{tabular}
    \label{table:auc_exploiter_reward}
\end{table}

 From Table \ref{table:auc_exploiter_reward}, we can also see that the adaptation metric and the robustness metric tend to be negatively correlated. To further analyze the relationship between adaptation (exploitation) and robustness (exploitability), we show the normalized area between curves (ABC) in Table \ref{table:exploitation_sensitivity_analysis}, which is the gap between the reward against the second exploiter opponent and the first exploiter opponent. Lower ABC indicates less sensitivity to opponent exploitation. This result, together with the result shown in Table \ref{table:auc_exploiter_reward}, verifies the well-known trade-off between exploitation and exploitability \cite{safe_exploit}: the settings with both the ensemble regularization term and the behavior cloning terms (reg\_bc\_rl and reg\_bc) are slightly more exploitable than their counterpart without the behavior cloning term (reg\_rl), which is an inevitable consequence of exploiting the interaction experience against the first exploiter opponent.

\begin{table}[t]
    \centering
    \caption{Normalized ABC (area between curves) between the second exploiter reward and the first exploiter reward. Lower ABC score indicates that the ego-agent's policy is less sensitive to opponent exploitation.}
    \begin{tabular}{lcr}
        \textbf{Swimmer} & & \\
        \hline
        Settings & Normalized ABC\\
        \hline
        Oracle opponent & 1.0\\
        Ensemble opponent & 0.0\\
        Reg\_bc\_rl & 0.27\\
        Bc\_rl & 1.23\\
        Reg\_rl & \textbf{0.05}\\
        Reg\_bc & 0.25\\
        \hline
    \end{tabular}
    
    \begin{tabular}{lcr}
        \\
        \textbf{Ant} & & \\
        \hline
        Settings & Normalized ABC\\
        \hline
        Oracle opponent & 1.0\\
        Ensemble opponent & 0.0\\
        Reg\_bc\_rl & 0.37\\
        Bc\_rl & 1.07\\
        Reg\_rl & \textbf{0.11}\\
        Reg\_bc & 0.58\\
        \hline
    \end{tabular}
    \label{table:exploitation_sensitivity_analysis}
\end{table}

\section{Conclusions}
This paper investigates safe adaptation which is an important problem in competitive MARL. In contrast to the widely-studied fast adaptation problem, our focus is on maintaining low exploitability during the adaptation. Our key innovation is the derivation of a novel ensemble regularization term from a Bayesian formulation of the MARL objective function. We show empirically that our proposed approach is effective both at adaptation to a previously unseen opponent given experience from a few interaction episodes and at maintaining low exploitability against an adversarial opponent that actively exploits the weakness of the ego-agent. Our ablation study and analysis reveal the effect of each term in our proposed loss function, as well as verify the well-known trade-off between exploitation and exploitability. Our work contributes to an important step towards building reliable intelligent robots that are able to operate safely in competitive multiagent scenarios against ever-changing adversarial opponents. 

\balance

\bibliography{IEEEexample}
\bibliographystyle{IEEEtran}

\end{document}